\documentclass[letterpaper]{article} 
\usepackage{aaai23}  
\usepackage{times}  
\usepackage{helvet}  
\usepackage{courier}  
\usepackage[hyphens]{url}  
\usepackage{graphicx} 
\usepackage{bbding}
\usepackage{multirow}
\usepackage{natbib}  
\usepackage{caption} 
\usepackage{algorithm}
\usepackage{algorithmic}
\usepackage{amsmath}
\usepackage{threeparttable}
\usepackage{booktabs}
\usepackage{newfloat}
\usepackage{listings}
\DeclareCaptionStyle{ruled}{labelfont=normalfont,labelsep=colon,strut=off} 
\floatstyle{ruled}
\newfloat{listing}{tb}{lst}{}
\floatname{listing}{Listing}
\title{Pose-Oriented Transformer with Uncertainty-Guided Refinement \\for 2D-to-3D Human Pose Estimation}
\author{
Han Li\textsuperscript{\rm 1}, Bowen Shi\textsuperscript{\rm 1}, Wenrui Dai\textsuperscript{\rm 1}\thanks{Corresponding author}, Hongwei Zheng\textsuperscript{\rm 1}, Botao Wang\textsuperscript{\rm 2},\\
Yu Sun\textsuperscript{\rm 2}, Min Guo\textsuperscript{\rm 2}, Chenglin Li\textsuperscript{\rm 1}, Junni Zou\textsuperscript{\rm 1}, Hongkai Xiong\textsuperscript{\rm 1}}
\affiliations{
\textsuperscript{\rm 1}Shanghai Jiao Tong University, Shanghai, China\quad \textsuperscript{\rm 2}Qualcomm AI Research\thanks{Qualcomm AI Research is an initiative of Qualcomm Technologies, Inc. Datasets were downloaded and evaluated by Shanghai Jiao Tong University researchers.}, Shanghai, China\\
\{qingshi9974, sjtu\_shibowen, daiwenrui, 1424977324\}@sjtu.edu.cn, \{botaow, sunyu, mguo\}@qti.qualcomm.com,\\ \{lcl1985, zoujunni, xionghongkai\}@sjtu.edu.cn
}

\usepackage{bibentry}

\begin{document}

\maketitle

\begin{abstract}
There has been a recent surge of interest in introducing transformers to 3D human pose estimation (HPE) due to their powerful capabilities in modeling long-term dependencies. However, existing transformer-based methods treat body joints as equally important inputs and ignore the prior knowledge of human skeleton topology in the self-attention mechanism. To tackle this issue, in this paper, we propose a Pose-Oriented Transformer (POT) with uncertainty guided refinement for 3D HPE. Specifically, we first develop novel pose-oriented self-attention mechanism and distance-related position embedding for POT to explicitly exploit the human skeleton topology. The pose-oriented self-attention mechanism explicitly models the topological interactions between body joints, whereas the distance-related position embedding encodes the distance of joints to the root joint to distinguish groups of joints with different difficulties in regression. 
Furthermore, we present an Uncertainty-Guided Refinement Network (UGRN) to refine pose predictions from POT, especially for the difficult joints, by considering the estimated uncertainty of each joint with uncertainty-guided sampling strategy and self-attention mechanism. Extensive experiments demonstrate that our method significantly outperforms the state-of-the-art methods with reduced model parameters on 3D HPE benchmarks such as Human3.6M and MPI-INF-3DHP. 


\end{abstract}

\section{Introduction}
\label{sec:in}
3D human pose estimation (HPE) aims to obtain the 3D spatial coordinates of body joints from monocular images or videos. It has attracted extensive attention in a wide range of applications such as autonomous driving, augmented/virtual reality (AR/VR) and virtual avatar. The 2D-to-3D pipeline is prevailing in recent works~\cite{2017simple,zhao2019semantic,2019Exploiting,li2021hierarchical}, where 2D joint coordinates are taken as the inputs to directly regress the 3D pose target.
Despite its promising performance, the 2D-to-3D pipeline is restricted by depth ambiguity caused by the many-to-one mapping from multiple 3D poses to one same 2D projection. 

\begin{figure}[!t]
    \centering
    \includegraphics[width=1.1 in]{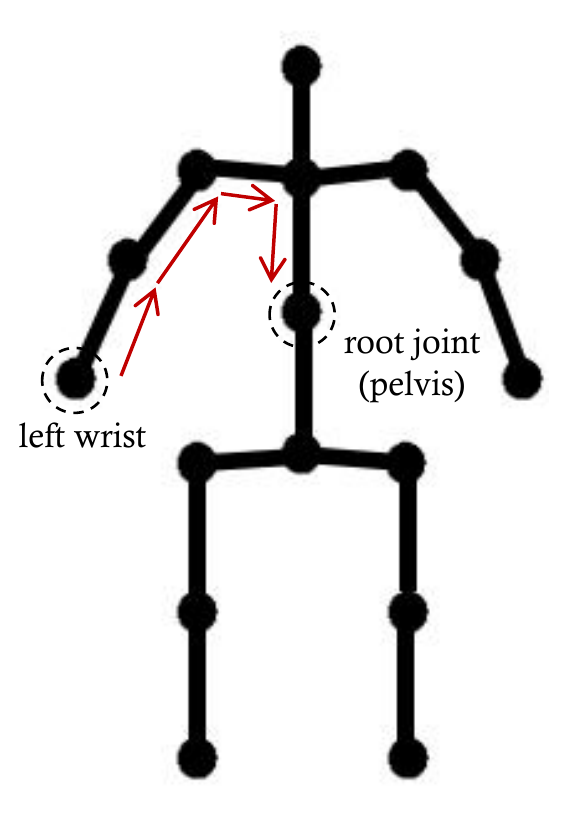} 
\includegraphics[width=2.1 in]{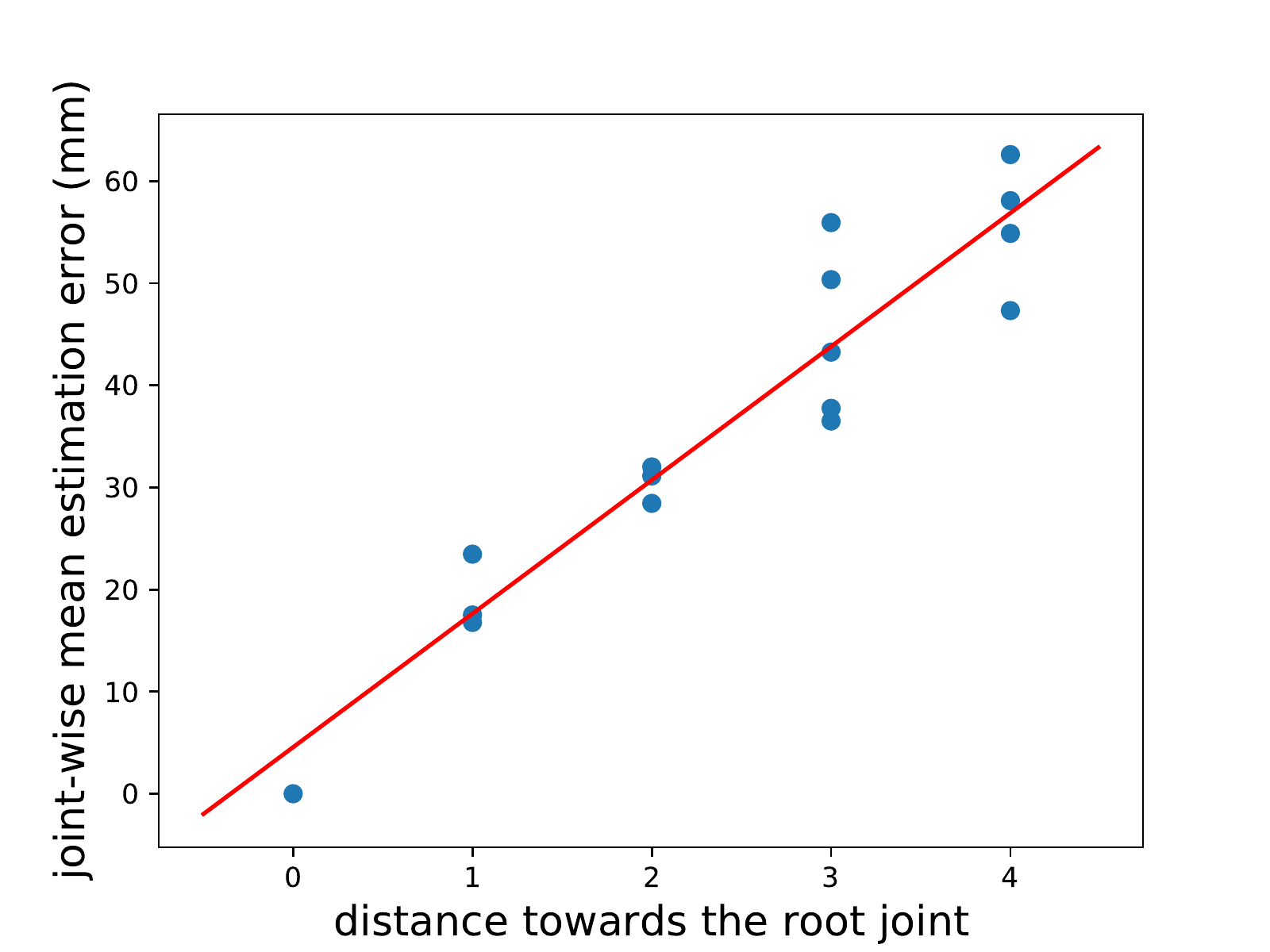}
    \caption{\textbf{Left: Human skeleton topology}. We consider the distance for each joint towards the root joint (pelvis) based on the human skeleton topology.
    \textbf{Right: Impact of distance towards the root joint on the joint-wise estimation error.} Based on a baseline model, we empirically find that joints far from the root joint tend to have large prediction errors. This inspires us to introduce targeted designs for these joints.}
    \label{fig:illustation_group}
\end{figure}

Considering that the human body can be modeled as a highly structured graph, the problem of depth ambiguity can be alleviated by exploiting the interactions between body joints. Graph convolution networks (GCNs) have been naturally adopted to exploit these interactions~\cite{zhao2019semantic,2019Exploiting,li2021hierarchical}. However, GCNs are usually limited in receptive fields and impede the relationship modeling. Inspired by the success of Transformer~\cite{vaswani2017attention}, the self-attention mechanism is leveraged in recent works~\cite{zheng20213d,zhu2021posegtac,zhao2022graformer,zhang2022mixste} to facilitate global interactions for 3D HPE and yield state-of-the-art performance.
\textbf{However, these methods treat body joints as input tokens of equal importance but ignore the human body priors (\emph{e.g.}, human skeleton topology) in designing the self-attention mechanism.} 



In this paper, we argue that introducing pose-oriented designs to the transformer is important for 3D HPE and thereby propose a Pose-Oriented Transformer (POT) for reliable pose prediction. We design a novel pose-oriented self-attention (PO-SA) mechanism for POT that is the first to explicitly exploit human skeleton topology without implicitly injecting graph convolutions. The relative distance is computed for each joints pair and is encoded as attention bias into the self-attention mechanism to enhance the ability of modeling the human skeleton dependence.
Furthermore, as shown in Figure~\ref{fig:illustation_group}, we empirically find that joints far from the root joint (pelvis) tend to have large prediction errors. To better model these difficult joints, we split body joints into several groups according to their distance toward the root joint and assign additional distance-related position embeddings to different groups.


In addition to POT, a second stage of pose refinement is developed to further improve the prediction of difficult joints. Specifically, we propose a transformer-based Uncertainty-Guided Refinement Network (UGRN) for pose refinement by explicitly considering the prediction uncertainty. The proposed UGRN comprises an uncertainty-guided sampling strategy and an uncertainty-guided self-attention (UG-SA) mechanism. The uncertainty-guided sampling strategy incorporates the estimated uncertainty for each joint (that implies the difficulty of prediction) into the learning procedure. The joint coordinates are sampled around the prediction from POT following a Gaussian distribution with the estimated uncertainty as variance. 
Then, we use the sampled coordinates as the input of UGRN to make the model more robust to errors. 
Subsequently, the UG-SA is developed in UGRN to reduce the contribution of the joints with high uncertainty during learning. 


This paper makes the following contributions:
\begin{itemize}
\item We propose a novel pose-oriented transformer for 3D HPE with the self-attention and position embedding mechanisms explicitly designed to exploit human skeleton topology. 
\item We present an uncertainty-guided refinement network to further improve pose predictions for difficult joints with uncertainty-guided sampling strategy and self-attention mechanism. 
\item We demonstrate our method achieves SOTA performance on the Human3.6M and  MPI-INF-3DHP benchmarks and shed light on the task-oriented transformer design for single-frame input human pose estimation. 
\end{itemize}

\section{Related Work}
\subsection{3D Human Pose Estimation} The methods of  3D human pose estimation can be divided into two categorizes: one-stage methods and two-stage methods. The one-stage methods take RGB image as input and directly predict the 3D pose.
Thanks to the development of deep learning, 
recent works~\cite{zhou2017towards,shi2020tiny,2018Ordinal,moon2019camera,lin2020hdnet,sun2017compositional} can leverage the  advantages of Convolutional Neural Networks (CNNs) to obtain promising results for  image-to-3D human pose estimation. In which
\cite{zhou2017towards} built a weakly-supervised transfer learning framework to make full use of mixed 2D and 3D labels, and  augmented the 2D pose estimation sub-network with a 3D depth regression sub-network to estimate the depth.  \cite{2018Ordinal} represented the space around the human body discretely as voxel and used 3D heatmaps to regress 3D human pose. Taking the feature extracted by CNNs as input, \cite{lin2021mesh} further proposed a graph-convolution-reinforced transformer to predict 3D pose. \cite{wehrbein2021probabilistic} proposed a normalizing flow method that can generate a diverse set of feasible 3D poses.

The second category of methods first estimate the 2D position of human joints from the input image, and then regress the 3D pose in the camera coordinate system. Pioneering work~\cite{2017simple} revealed that only using 2D joints as input can also gets highly accurate results, and proposed a simple yet effective baseline for 3D HPE. Since the human body can be regarded as a highly structured graph, \cite{zhao2019semantic} proposed Semantic Graph Convolution (SemGConv) for 3D HPE, it added a parameter matrix to learn the semantic relations among body joints. \cite{zou2020high} further extended SemGConv to a high-order GCN To learn long-range dependencies among body joints .  Nevertheless, GCN-based methods still suffer from limited receptive field. In this work, we leverage the powerful long-term modeling capability of transformer to  construct our model.

\subsection{Transformer and Self-Attention Mechanism}
Transformer was firstly introduced in~\cite{vaswani2017attention} for the natural language processing (NLP) tasks such as machine  translation, whose core component is the self-attention mechanism that can model the long-term dependence of the input sequential data. Recently, with the appearance of VIT~\cite{dosovitskiy2020image}, transformer also attracted much attention in various visual tasks. In addition, \cite{ying2021transformers} also generalized transformer to graph-structured data for graph-level predictions tasks including link prediction and knowledge graphs. For the 3D 
HPE, PoseFormer~\cite{zheng20213d} first built a  transformer-based model to sequentially capture the temporal and spatial dependency of the input 2D pose sequence. PoseGTAC~\cite{zhu2021posegtac} and Graformer~\cite{zhao2022graformer} both injected graph convolution into transformer in different ways to exploit the structure information of human skeleton topology. However, we argue that simply stacking self-attention and graph convolution can not fully utilize the human skeleton topology and propose our pose-oriented transformer to take the  topology information  into account in the self-attention mechanism.

\begin{figure*}[!t]
\renewcommand{\baselinestretch}{1.0}
\centering
\includegraphics[scale=0.73]{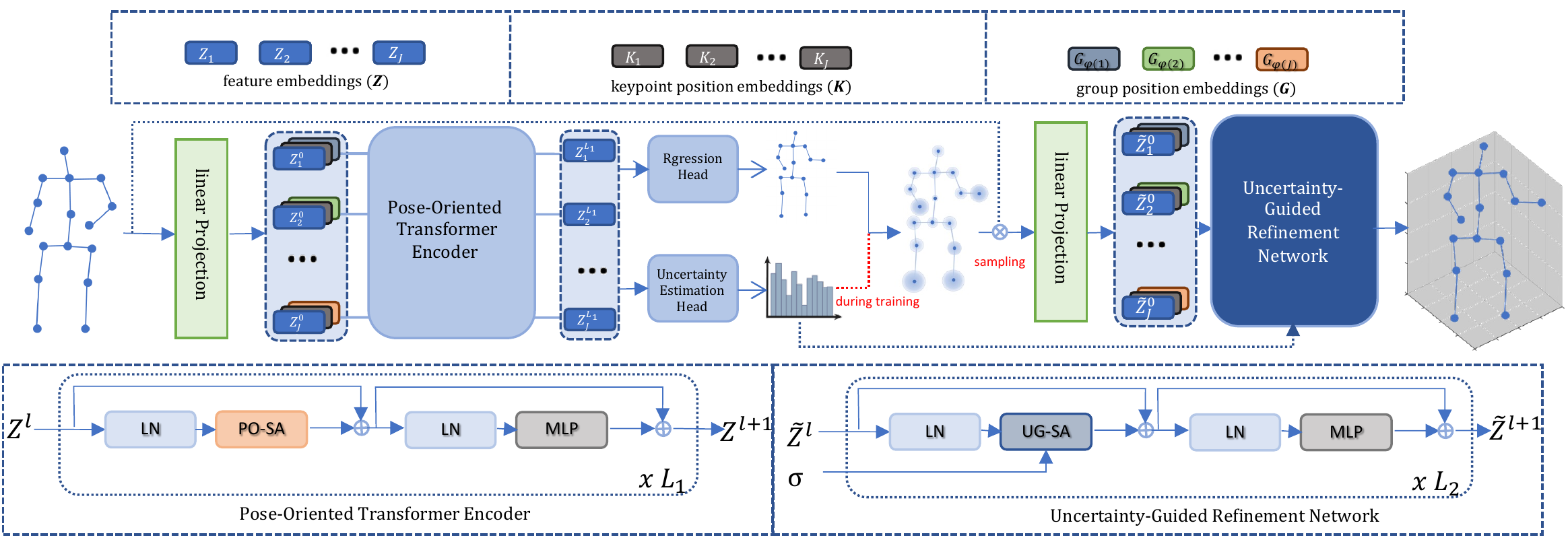}
\caption{The overview of proposed method, which contains two major module: pose-oriented transformer (POT) and uncertainty-guided refinement network (UGRN). Given the 2D pose $X\in R^{J\times 2}$ estimated by an off-the-shelf 2D pose detector,  POT with pose-oriented attention and position embedding designs are first used for pose-related feature extracting and first-stage 3D pose predicting. Then, UGRN leverage uncertainty information $\sigma \in R^{J\times 3}$ to  generate refined pose $\hat{Y}\in R^{J\times3}$. }\label{fig:overview}
\end{figure*}

\subsection{Uncertainty Estimation}
Uncertainty in the deep learning models can be categorized into two types: aleatoric uncertainty and epistemic uncertainty. It can be estimated by sampling-based method~\cite{glorot2010understanding} and dropout method~\cite{gal2016dropout}.  \cite{kendall2017uncertainties} further revealed that the heteroscedastic uncertainty dependent on the input data is vitally important for computer vision application. For example, \cite{song2021uncertain} considered the uncertainty of the noisy input data and proposed the uncertain graph neural networks for facial action unit detection.
\cite{wang2021data} utilized the data-uncertainty as guidance to propose a multi-phase learning method for semi-supervised object detection.
\cite{yang2021uncertainty}  combined the benefits of Bayesian learning and transformer-based reasoning, and built an uncertainty-guided transformer for camouflaged object detection.
However, previous 2D-to-3D HPE methods did not take uncertainty information of human pose into account in the training and inference procedure.
For our work, we estimate the uncertainty for each joint of first-stage 3D pose and propose our UG-sampling and UG-SA to obtain the refined 3D pose.

\section{Method}
The overview of the proposed method is depicted in Figure~\ref{fig:overview}. Our method is a two-stage framework which consists of two major module: pose-oriented transformer (POT) and uncertainty-guided refinement network (UGRN). Given the 2D pose $X\in R^{J\times 2}$ estimated by an off-the-shelf 2D pose detector from an image,  POT is designed by utilizing human skeleton topology for better pose-related feature extracting and first-stage 3D pose predicting, while UGRN leverages uncertainty information $\sigma \in R^{J\times 3}$ to further refine the predicting pose. Details are included in the following.

\subsection{Preliminaries}
In this work, we leverage transformer to model the long-distance relationship between body joints. We first briefly introduce the basic components in the transformer, including multi-head self-attention (MH-SA), position-wise feed-forward network (FFN) and position embeddings.
    \subsubsection{MH-SA} The basic self-attention mechanism transfers the inputs $Z\in R^{N\times C}$ into corresponding $query~ Q$, $key~ K$ and $value~ V$ with the same dimensions $N\times C$ by projection matrices $P^Q,P^K,P^V\in R^{C\times C}$ respectively, where $N$ denotes the sequence length, and $C$ is the number of hidden dimension.
\begin{equation}
    Q=ZP^Q, ~~ K=ZP^K, ~~ V=ZP^V,
\end{equation}
Then we can calculate self-attention by:
\begin{equation}
\label{eq:MH-SA}
    A = QK^T/\sqrt{d}, ~~\textit{MH-SA}(X) = {softmax}(A)V,
\end{equation}
where $A\in R^{N\times N }$ denotes the attention weight matrix. Based on the basic self-attention, MH-SA further splits the $Q,K,V$ for $h$ times to perform attention in parallel and then the outputs of all the heads are concatenated. 
\subsubsection{FFN} position-wise FFN is used for non-linear feature transformation and it contains two Multilayer Perceptron (MLP) and an GELU activation layer. This procedure can be formulated as follows:
\begin{equation}
    FFN(X) = MLP(GELU(MLP(X)))+X.
\end{equation}

\subsubsection{Position Embeddings}
As  MH-SA and FFN in transformer are
permutation equivariant operation, additional mechanisms are required to encode the structure of input data into model. In particular, we can utilize sine and cosine functions or learnable vectors as the position embeddings, which can be formulated as
\begin{equation}
    P_t =PE(t)\in R^{C}, 
\end{equation}
where $t$ denotes the position index.

\subsection{Pose-oriented Transformer}
POT aims at better utilizing the human skeleton information for feature extracting. It includes target position embedding and self-attention design for 3D HPE. Specifically, given the input 2D joints $X\in R^{J\times2}$, we first project it into high-dimensional feature embeddings $Z\in R^{J\times C}$, where $J$ denotes the number of human body joints and $C$ denotes the embedding dimension. Then we add keypoint position embeddings $K$ and our proposed group position embeddings $G$ to $Z$ as the input of POT encoder. In POT encoder, we also design pose-oriented self-attention (PO-SA) which takes the topological connections of body joints into consideration. 

\subsubsection{Keypoint and Group Position Embeddings}

\begin{figure}[!t]
\renewcommand{\baselinestretch}{1.0}
\centering
\includegraphics[scale=0.3]{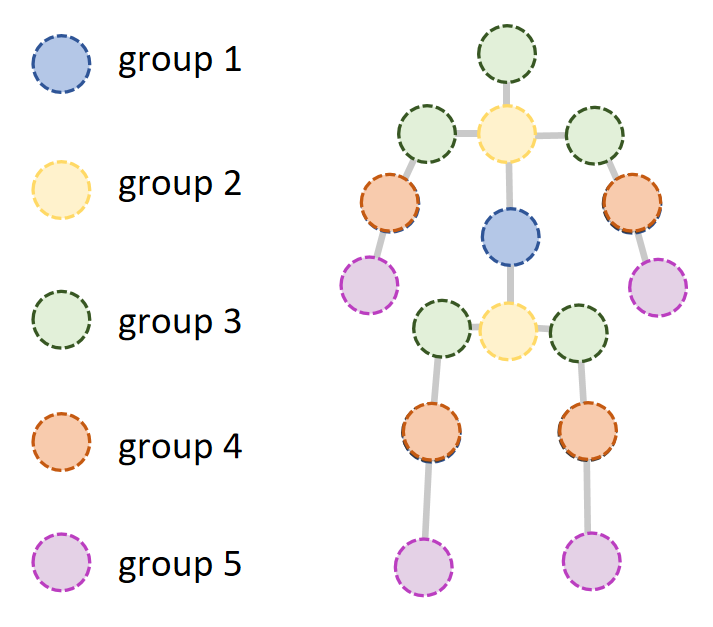}
\caption{The depiction of distance-related group for human body joints.}
    \label{fig:group}
\end{figure}

Following previous design~\cite{zheng20213d,zhang2022mixste}, we first introduce a learnable keypoint position embeddings $K \in R^{J\times C}$ to represent the absolute position of each body joint. In addition, as shown in Figure~\ref{fig:group}, according to the distance between each joint and the root joint (Pelvis), we split body joints into five groups and design another learnable embeddings called group position embeddings, \textit{i.e.} 
, $G \in R^{5\times C}$. Therefore, additional distance-related knowleage can be encoded into model, helping transformer better model the difficult body joints that are far from the root. In this way, the input of pose-oriented transformer encoder, $Z^{(0)}$, can be obtained by:
\begin{equation}
    Z_i^{(0)} = Z_i+K_i +G_{\varphi(i)}, ~~ for~~ i\in [1,\cdots,J],
\end{equation}
 where $i$ is the joint index and $\varphi(i)=\mathcal{D}(i,1)$ represents the shortest path distance between $i$-th joint and the root joint. 


\subsubsection{Pose-Oriented Self-Attention (PO-SA)} We also propose our pose-oriented self-attention (PO-SA) that explicitly modeling the topological connections of body joints. Specifically, we compute the relative distance for each joints pair $(i,j)$, and encode it as the attention bias for the self-attention mechanism. In this way, we rewrite the self-attention in Eq~(\ref{eq:MH-SA}), in which the $(i,j)$-th element of attention matrix $A$ can be computed by:
\begin{equation}
\label{eq:PO-SA}
    A_{i,j} = (Z_iP^Q)(Z_jP^K)^T/\sqrt{d} +\Phi(\mathcal{D}(i,j)),
\end{equation}
where $\Phi$ is a MLP network which projects the relative distance (1-dimension) to an H-dimension vector where H is the number of heads in the SA mechanism, it makes each PO-SA have the ability to adjust the desired distance-related receptive field and the additional parameters can be ignored.  

\subsubsection{POT Encoder}
Based on the PO-SA, we can obtain output features by sending $Z^{(0)}$ to a cascaded transformer with $L_1$ layers. These procedure can be formulated as :
\begin{align}
Z^{\prime l} &= \textit{PO-SA}(\textit{LN}(Z^{l-1}))+Z^{l-1},\\
Z^{l} &= \textit{FFN}(\textit{LN}(Z^{\prime l}))+Z^{\prime l},
\end{align}
where $LN(\cdot)$ represents the layer normalization and  $l\in\left[1,2,\cdots,L_1\right]$ is the index of POT encoder layers. 

\subsubsection{Regression Head}
In the regression head, we apply a MLP on the output feature $Z^{L_1}$ to perform pose regression, generating the first-stage 3D pose $\widetilde{Y}\in R^{J\times 3}$.

\subsection{Uncertainty-guided Refinement}
Taking the first-stage 3D pose $\widetilde{Y}$ from POT, we further send it together with the input 2D pose $X$ to another Uncertainty-guided Refinement Network (UGRN) for pose refinement. The proposed UGRN contains the following components.  

\subsubsection{Uncertainty Estimation }
We first model the uncertainty for each joint. Specifically, features of POT encoder $Z^{L_1}$ are sent to another uncertainty estimation head, producing the uncertainty $\sigma\in R^{J\times3}$ of the first-stage 3D poses by using an uncertainty estimation loss $\mathcal{L}_{\sigma}$~\cite{kendall2017uncertainties}. 


\subsubsection{Uncertainty-Guided Sampling}


Instead of directly utilizing the first-stage 3D predictions $\widetilde{Y}$, we randomly sample 3D coordinates $\bar{Y}$ around $\widetilde{Y}$ according to a Gaussian distribution $\mathcal{N}(\widetilde{Y},\sigma)$ with the predicted uncertainty $\sigma$ as variance, and send the sampled coordinates to UGRN. This uncertainty-guided sampling strategy ensures that the sampled coordinates have large variance on difficult joints, which requires the model to focus more on making use of context from other joints to compensate for the difficult joint predictions, thus further enhancing the model robustness. 

To enable correct back-propagation, we employ a re-parameterization trick to draw a sample $\epsilon$ from the standard Gaussian distribution $\mathcal{N}(\textbf{0},\textbf{1})$ randomly, $i.e.$, $\epsilon \sim\mathcal{N}(\textbf{0},\textbf{1})$. In this way, we can obtain the sampled 3D coordinates by:
\begin{equation}
    \bar{Y} = \widetilde{Y} +\sigma\cdot\epsilon.
\end{equation}

Note that this sample strategy is only implemented in the training stage. In the inference stage, we set $\bar{Y} = \widetilde{Y}$ directly.

\subsubsection{Uncertainty-guided Refinement Network}
After obtaining the sampled 3D pose $\bar{Y}$, we first concatenate it with the input 2D pose $X$ and obtain $\widetilde{X}$, \textit{i.e.}, $\widetilde{X} = \textit{Concat}(\bar{Y},X)$. Then we project $\widetilde{X}$ to feature embeddings $\widetilde{Z}$ and equip them with keypoint position embeddings $K$ and group position embedding $G$:
\begin{equation}
    \widetilde{Z}_i^{(0)} = \widetilde{Z}_i+K_i +G_{\varphi(i)}, ~~ for~~ i\in [1,J].
\end{equation}

Next, $\tilde{Z}_i^{(0)}$ is sent to the following $L_2$ transformer layers of UGRN to perform uncertainty-guided refinement. The transform layers of UGRN is similar to those of POT, but we replace the distance-related term of Eq.~\ref{eq:PO-SA} with uncertainty guildance to dynamically adjust the attention weights:

\begin{equation}
    A_{i,j} = (Z_iP^Q)(Z_jP^K)^T/\left(\sqrt{d}\cdot\textit{Sum}(\sigma_j)\right),
\end{equation}
where $\sigma_j\in R^{3}$ is the predicted uncertainty of $j$-th joint. The above uncertainty-guided self-attention (UG-SA) ensures that the body joints with high uncertainty will contribute less in the self-attention mechanism, which can not only alleviate the error propagation, but also enhance the context understanding ability of the model.

Finally, we apply another regression head to $\tilde{Z}^{L2}$ and generate our second-stage refined 3D pose $\hat{Y}\in R^{J\times 3}$.

\begin{table*}[!t]
\setlength{\tabcolsep}{0.5pt}
\centering
\caption{Quantitative evaluation results using MPJPE in millimeter on Human3.6M . No rigid alignment or transform is applied in post-processing. We  split this table  into 2 groups. The inputs for the top group methods are the detection 2D pose, SH denotes the 2D pose detected by Stacked Hourglass network~\cite{newell2016stacked}, and CPN  denotes the cascaded pyramid network~\cite{chen2018cascaded}. The inputs for the bottom group are ground truth (GT)  of 2D pose. Best results
are showed in bold.}\label{table:MPJPE}	
\begin{tabular}{l|ccccccccccccccc|c}
\hline
Methods &Dire.&Disc.&Eat&Greet&Phone&Photo&Pose&Puch.&Sit&SitD.&Smoke&Wait&WalkD&Walk&WalkT&\textbf{Avg.} \\  
\hline\hline
\cite{2017simple} (SH) &51.8&56.2&58.1&59.0&69.5&78.4&55.2&58.1&74.0&94.6&62.3&59.1&65.1&49.5&52.4&62.9 \\ 
\cite{zhao2019semantic} (SH) &48.2&60.8&51.8&64.0&64.6&{\bf53.6}&51.1&67.4&88.7&{\bf57.7}&73.2&65.6&{\bf48.9}&64.8&51.9&60.8\\
\cite{liu2020comprehensive} (CPN) &46.3 &52.2& 47.3& 50.7& 55.5 &67.1& 49.2&{\bf46.0}& 60.4 &71.1& 51.5& 50.1& 54.5 &40.3& 43.7& 52.4\\
\cite{zou2020high}(CPN) &49.0& 54.5& 52.3& 53.6 &59.2& 71.6& 49.6 &49.8& 66.0 &75.5 &55.1 &53.8& 58.5& 40.9 &45.4 &55.6 \\
\cite{2021Graph}(CPN) &{\bf 45.2}&  {\bf49.9}& 47.5& 50.9& 54.9&  66.1&   48.5&  46.3&  59.7&  71.5&51.4&{\bf48.6}&53.9& {\bf 39.9}& 44.1& 51.9 \\ 
\hline
Ours (CPN)&47.9&50.0&47.1&51.3&{\bf51.2}&59.5&\bf{48.7}&46.9&\bf{56.0}&61.9&{\bf51.1}&48.9&54.3&40.0&{\bf42.9}&{\bf 50.5} \\ 
\hline\hline
\cite{2017simple} (GT)& 37.7 & 44.4 & 40.3 & 42.1 & 48.2 & 54.9 & 44.4 & 42.1 & 54.6 & 58.0 & 45.1 & 46.4 & 47.6 & 36.4 & 40.4 & 45.5 \\
\cite{zhao2019semantic} (GT) & 37.8 & 49.4 & 37.6 & 40.9 & 45.1 &41.4& 40.1 & 48.3 & 50.1 & 42.2& 53.5 & 44.3 & 40.5 & 47.3 & 39.0 & 43.8 \\
\cite{liu2020comprehensive} (GT) & 36.8 & 40.3 & 33.0 & 36.3 & 37.5 & 45.0 & 39.7 & 34.9 & 40.3 & 47.7 & 37.4 & 38.5 & 38.6 & 29.6 & 32.0 & 37.8 \\
\cite{2021Graph}  (GT) &  35.8 & 38.1 & 31.0 & 35.3 & 35.8 & 43.2 & 37.3 & 31.7 & 38.4 & 45.5 & 35.4 & 36.7 & 36.8 & 27.9 & 30.7 & 35.8\\
\cite{zhao2022graformer} (GT) & \textbf{32.0} & \textbf{38.0} & 30.4 & 34.4 & \textbf{34.7} & 43.3 & \textbf{35.2} & \textbf{31.4} & 38.0& 46.2 & 34.2 & 35.7 & 36.1 & 27.4 & 30.6 & 35.2\\
\hline
Ours (GT)&32.9&38.3&{\bf28.3}&{\bf33.8}&34.9&{\bf38.7}&37.2&{\bf30.7}&{\bf34.5}&{\bf39.7}&{\bf33.9}&{\bf34.7}&{\bf34.3}&{\bf26.1}&{\bf28.9}&{\bf33.8} \\ 
\hline
\end{tabular}
\end{table*}

\begin{table*}[!t]
\centering
\caption{Results on the test set of MPI-INF-3DHP~\cite{mehta2017monocular}  by scene.  The results are shown in PCK and AUC.}\label{table:MPI} 
\begin{tabular}{l|c |ccc|cc}
\hline
Methods &Trainning data &~GS~& noGS &Outdoor &ALL (PCK~$\uparrow$) &ALL (AUC~$\uparrow$)\\ \hline \hline
\cite{2017simple}&H36M &49.8&42.5&31.2&42.5&17.0\\
\cite{mehta2017monocular}&H36M&70.8&62.3&58.8&64.7&31.7\\
\cite{yang20183d} &H36M+MPII&-&-&-&69.0&32.0\\
\cite{zhou2017towards} &H36M+MPII&71.1&64.7&72.7&69.2&32.5\\
\cite{luo2018orinet}  &H36M&71.3&59.4&65.7&65.6&33.2\\
\cite{ci2019optimizing}&H36M&74.8&70.8&77.3&74.0&36.7\\
\cite{zhou2019hemlets} &H36M+MPII &75.6&71.3&80.3&75.3&38.0\\
\cite{2021Graph}    &H36M &81.5&81.7&75.2&80.1&45.8\\
\cite{zhao2022graformer} &H36M& 80.1&77.9&74.1&79.0&43.8\\ \hline
Ours &H36M& \textbf{86.2}&\textbf{84.7}&\textbf{81.9}&\textbf{84.1}&\textbf{53.7}\\ \hline
\end{tabular}
\end{table*}

\subsection{Loss Function}
\subsubsection{Stage I} We first train our POT for the first-stage 3D pose regressing. The objective function can be formulated as :
\begin{equation}
\mathcal{L}_\mathrm{stageI} =\frac{1}{J}\sum_{i=1}^J{\left( \left\| \widetilde{Y}_{i}-Y_{i} \right\| ^2 \right)},
\end{equation}
where $\widetilde{Y}_i$ and $Y_i$ are the estimated first-stage 3D positions and the ground truth of $i$-th joint respectively.
\subsubsection{Stage II} We aim to predict the uncertainty correctly as well as estimate an accurate refined 3D pose in Stage II. During this stage, we freeze the model parameters of POT and only train the UGRN for stable results. Following~\cite{kendall2017uncertainties}, we set our uncertainty estimation loss as:
\begin{equation}
\mathcal{L}_{\sigma} =    \frac{1}{J}\sum_{i=1}^J{\left( \left\| \frac{\widetilde{Y}_{i}-Y_{i}}{\sigma_i} \right\| ^2+\log(\|\sigma_i\|^2) \right)}.
\end{equation}
In addition, we also apply L2 loss to minimize the errors between the refined 3D poses and ground truths:
\begin{equation}
\mathcal{L}_\mathrm{refine} = \frac{1}{J}\sum_{i=1}^J{\left(\left\|\hat{Y}_{i}-Y_{i}\right\|^2\right)},
\end{equation}
The final loss function of Stage II is computed by  $ \mathcal{L}_\mathrm{stageII} = \mathcal{L}_\mathrm{refine} +\lambda\mathcal{L}_{\sigma}$, where
$\lambda$ is the trade-off factor. We set $\lambda$ to 0.001 such that the two loss terms are of the same order of magnitude.

\section{Experiments}
\label{sec:exp}
\subsection{Experimental Setups}
\subsubsection{Dataset}  Human3.6M dataset~\cite{ionescu2013human3} is widely used in the 3D HPE task which provides 3.6 million indoor RGB images, including 11 subjects actors performing 15 different actions.
For fairness, we follow previous works~\cite{2017simple,zhao2019semantic,2021Graph} and take 5 subjects (S1, S5, S6, S7, S8) for training and the other 2 subjects (S9, S11) for testing. In our work, We evaluate our proposed method and conduct ablation study on the Human3.6M dataset.
Besides, the MPI-INF-3DHP~\cite{mehta2017monocular} test set  provides images in three different scenarios: studio with a green screen (GS), studio without green screen (noGS) and outdoor scene (Outdoor). 
We also apply our method to it to demonstrate the generalization capabilities of our proposed method.


\subsubsection{Evaluation metrics} 
For Human3.6M, we follow previous works~\cite{2017simple,zhao2019semantic} to use the mean per-joint position error (MPJPE) as evaluation metric. MPJPE computes the per-joints mean Euclidean distance between the predicted 3D joints and the ground truth after the origin (pelvis) alignment. For MPI-INF-3DHP, we employ 3D-PCK and AUC as evaluation metrics.


\subsubsection{Implement details}
In our experiment, we set the dimension of embeddings to 96 and adopts 6 heads for self-attention with a dropout rate of 0.25. The MLP ratio of FFN is set to 1.5 to reduce the model parameters.
We implement our method within the PyTorch framework. During the training stage, we adopt the Adam ~\cite{kingma2014adam} optimizer. For both Stage I and Stage II, the learning rate is initialized to 0.001 and decayed by 0.96 per 4 epochs, and we train each stage for 25 epochs using a mini-batch size of 256. We initialize weights of the our model using the initialization method described in~\cite{glorot2010understanding}. We also adopt Max-norm regularization to avoid overfitting.

\subsection{Comparison With the State-of-the-Art}
The performance compared with the state-of-the-art are shown in Table~\ref{table:MPJPE}. In the top group, following the setting of previous works \cite{pavllo20193d,zhou2017towards,2019Exploiting},
We use the cascaded pyramid network (CPN)~\cite{chen2018cascaded} as 2D pose detector to obtain 2D joints for benchmark evaluation. 
In the bottom group, we take the ground truth (GT) 2D pose as input to predict the 3D human pose. It can be seen that,  our method outperforms all other methods with both GT and detected 2D pose as input, demonstrating the effectiveness of our method. 

\subsection{Generalization Ability}  
We further apply our model to MPI-INF-3DHP to test the generalization abilities.  As shown in Table~\ref{table:MPI}, our model achieves 84.1 in PCK and 53.7 in AUC while only using  Human3.6M dataset for training, which outperforms all the previous SOTA by a large margin. These results verify the strong generalization capability of our method.

\begin{table}[!t]
\renewcommand{\baselinestretch}{1.0}
\renewcommand{\arraystretch}{1.0}
\setlength{\tabcolsep}{5pt}
\setlength{\abovecaptionskip}{3pt}
\centering
\caption{Ablation Study on different pose-oriented design in the pose-oriented transformer. }\label{table:pose-oriented-design}
\begin{tabular}{ccc|c|c}
\hline
 \multicolumn{2}{c}{\textbf{position embeddings}} &\multirow{2}{*}{PO-SA}&\multirow{2}{*}{MPJPE(mm)}&\multirow{2}{*}{\#Param} \\
\cline{1-2} 
keypoint& group &&& \\ \hline
\Checkmark&&&37.57&0.97M\\ \hline
\Checkmark&\Checkmark&&36.69&0.97M\\ \hline
\Checkmark&&\Checkmark &36.43&0.98M \\ \hline
\Checkmark&\Checkmark&\Checkmark&\textbf{35.59}&0.98M\\ 
\hline
\end{tabular}
\end{table}
\begin{table}[!t]
\renewcommand{\baselinestretch}{1.0}
\renewcommand{\arraystretch}{1.0}
\setlength{\abovecaptionskip}{3pt}
\centering
\caption{Ablation Study on Uncertainty-Guided Refinement.}\label{table:UGRN}
\begin{tabular}{l|c|c}
\hline
Method &MPJPE(mm)&\#Param  \\\hline
POT & 35.59 &0.79M \\ \hline
POT+UGRN & 34.72 & 0.98M \\ \hline 
POT+UGRN+UG-Sampling & \textbf{33.82} & 0.98M \\ \hline
\end{tabular}
\end{table}
\begin{table}[!t]
\renewcommand{\baselinestretch}{1.0}
\renewcommand{\arraystretch}{1.0}
\setlength{\abovecaptionskip}{3pt}
\centering
\caption{Ablation study on UG-SA}\label{table:UG-SA}
\begin{tabular}{l|c|c}
\hline
Method &MPJPE(mm)&\#Param  \\\hline
POT+UGRN (MH-SA) & 35.22 & 0.98M \\ \hline 
POT+UGRN (PO-SA) & 35.07 & 0.98M \\ \hline 
POT+UGRN (UG-SA) & \textbf{34.72} & 0.98M \\ \hline 
\end{tabular}
\end{table}
\begin{table}[!t]
\renewcommand{\baselinestretch}{1.0}
\renewcommand{\arraystretch}{1.0}
\setlength{\abovecaptionskip}{3pt}
\centering
\caption{Ablations on different parameters of POT and UGRN. $L_1$ and $L_2$ are the number of layers of POT encoder and UGRN, respectively. $C$ is the embedding dimension.}\label{table:para-pot}
\begin{tabular}{ccc|c|c}
\hline
$L_1$& $L_2$& $C$  &MPJPE(mm)&\#Param  \\\hline
4 & 1 & 96 & 37.08 & 0.33M \\ \hline 
8 & 2 & 96  & 35.20 & 0.66M \\ \hline 
\textbf{12} & \textbf{3} & 96 & \textbf{33.82} & 0.98M \\ \hline 
16 & 4 & 96 & 34.47 & 1.31M \\ \hline  \hline
12 & 3 & 48  & 34.20 & 0.25M \\ \hline 
12 & 3 & \textbf{96} & \textbf{33.82}& 0.98M \\ \hline 
12 & 3 & 144 &  34.68 & 2.20M \\ \hline 
\end{tabular}
\end{table}
\begin{table}[!t] 
\renewcommand\baselinestretch{1.0}
\renewcommand\arraystretch{1.0}
\setlength{\abovecaptionskip}{3pt}
\centering
\caption{Comparison on model complexity.}\label{tab:model_complexity}
\resizebox{\columnwidth}{!}{
\begin{tabular}{l|c|c}
\hline
Method &MPJPE(mm)&\#Param  \\\hline
Pre-Aggre~\cite{liu2020comprehensive} & 37.80 & 4.22M \\ \hline
Graph SH~\cite{2021Graph}  & 35.80 &3.70M \\ \hline
Modulated GCN~\cite{zou2021modulated} & 37.43 & 1.10M \\ \hline
Graformer~\cite{zhao2022graformer} & 35.20 & 0.62M \\ \hline \hline 
Our-S &34.20 &0.25M \\ \hline
Our-L &33.82 &0.98M \\ \hline
\end{tabular}}
\end{table}

\begin{figure}[!t]
\renewcommand{\baselinestretch}{1.0}
\centering
 \includegraphics[width=3. in]{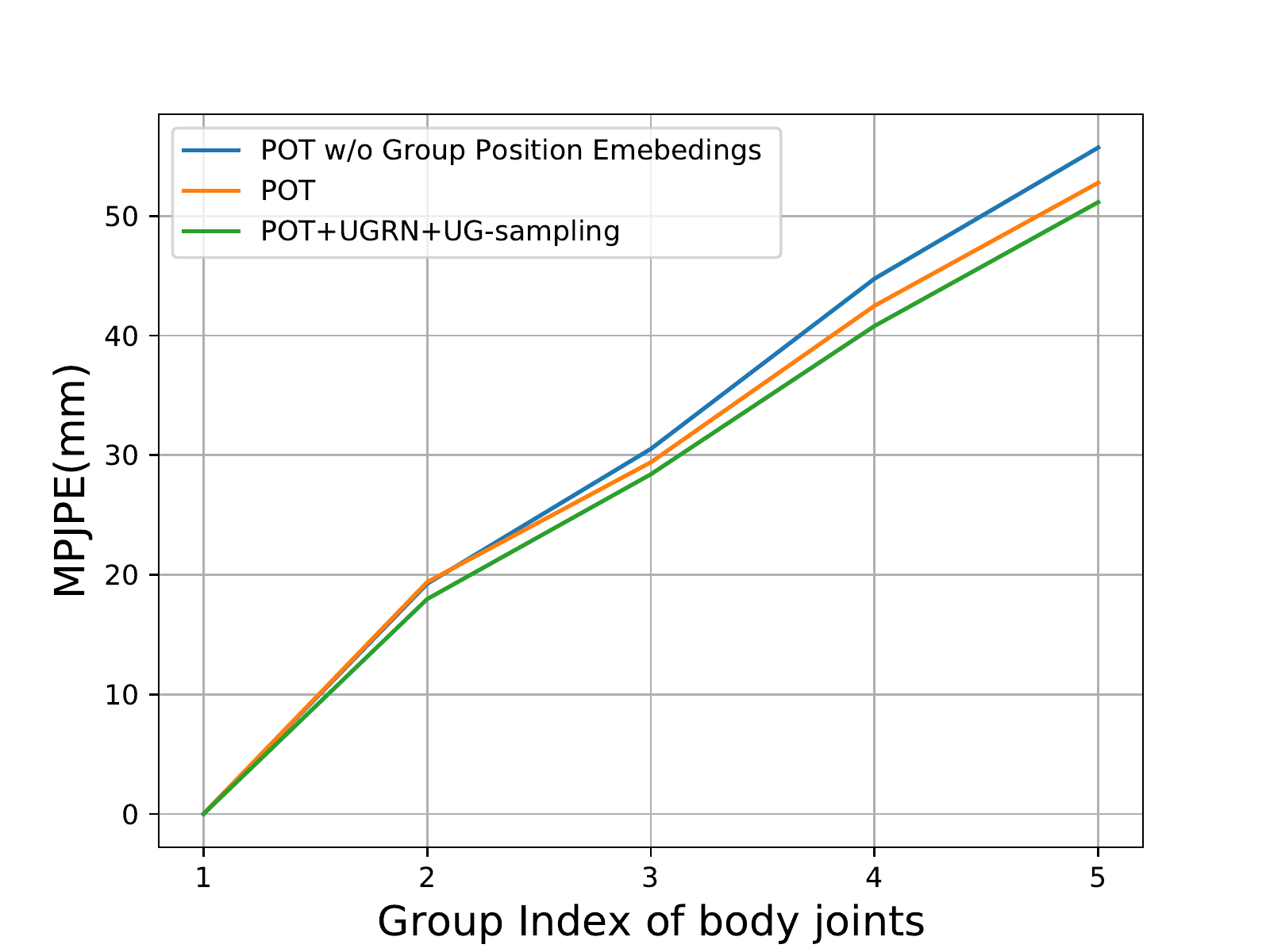}
\caption{Analysis on difficult joints. Our proposed group position embeddings and uncertainty-guided refinement mainly benefit  the difficult joints in group 4 and 5.}\label{fig:Hard joints}
\end{figure}

\begin{figure*}[!t]
\renewcommand{\baselinestretch}{1.0}
\centering 
\includegraphics[width=6.5 in]{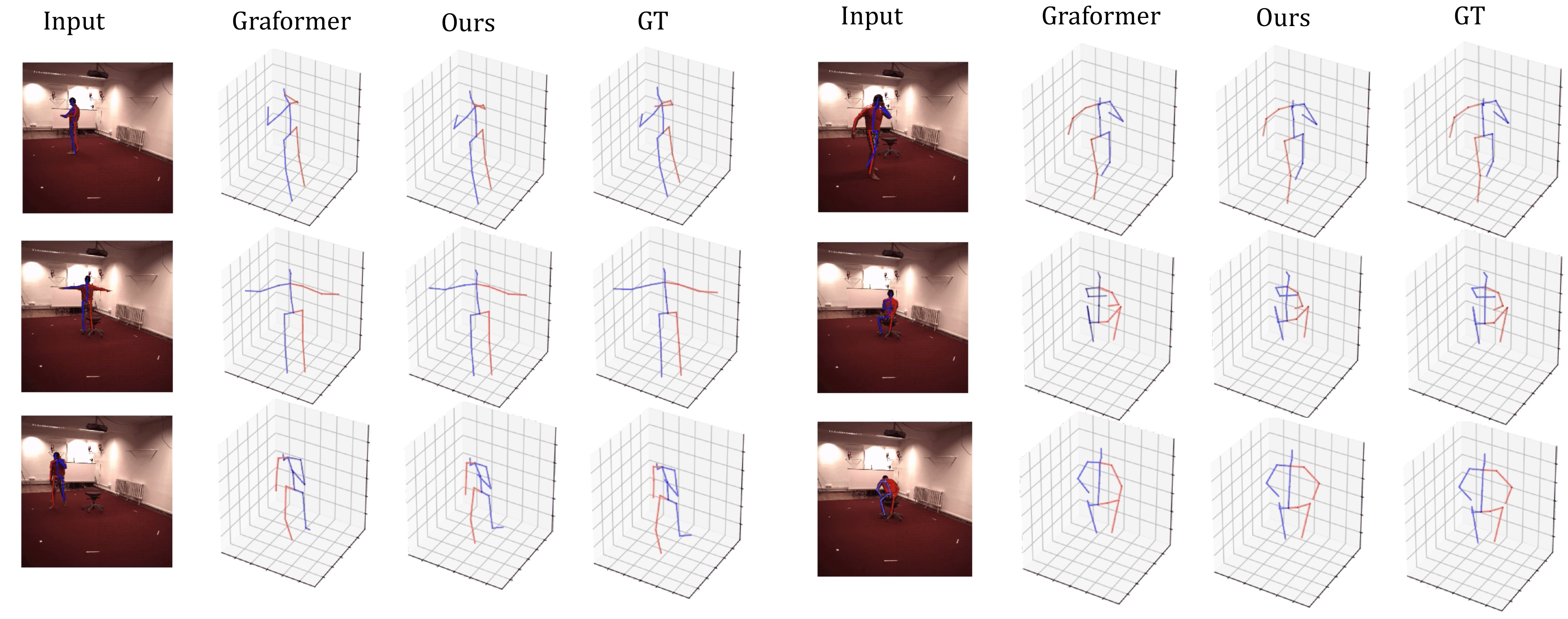}
\caption{Qualitative results  on Human3.6M.}\label{fig:vis}
\end{figure*}

\subsection{Ablation Study and Discussion}
We conduct a series of ablation studies to better understand how each component affects the performance. The 2D ground truth (GT) is taken as input in the ablation.
\subsubsection{Effect on different pose-oriented design}
We first diagnose how each pose-oriented design in the POT affects the performance.  In this section, the UGRN is excluded and the first stage 3D pose $\widetilde{Y}$ is used for evaluated. 
As shown in Table~\ref{table:pose-oriented-design}, our method achieves the best performance when all the pose-oriented designs are included. Compared with only using keypoint position embeddings , we achieve 0.88mm (37.57mm to 36.69mm) improvement by adding the distance-related group position embeddings, proving that the representation of difficult joints is effectively facilitated.
In addition, by replacing the standard self-attention with our PO-SA, we also achieve 1.10mm (36.69mm to 35.59mm) improvement with only 0.01M model parameters increase, which reflects the benefits of enhancing the ability of modeling the topological interactions. 

\subsubsection{Effect on uncertainty-guided refinement}
We then inspect how uncertainty-guided refinement benefits performance. It can be seen from Table~\ref{table:UGRN} that our first-stage prediction obtained directly by POT can achieve 35.59 mm in MPJPE, while adding UGRN for refinement can bring 0.83mm (35.59mm to 34.72mm) performance improvement, and UG-sampling can facilitate the learning procedure and further bring 0.9 mm (34.72mm to 33.82mm) gains. To demonstrate that the performance improvement is not brought by the increased model parameters, we also test other refinement model design using other kinds of self-attention, and the results are shown in Table~\ref{table:UG-SA}. When we replacing UG-RA with standard MH-SA, the performance degrades from 34.72mm to 35.22mm. In addition, when using the proposed PO-SA in the UGRN, the performance also degrades (34.72mm to 35.07mm), which reflects that the uncertainty-related information is more important than distance-related information in the second refinement stage.

\subsubsection{Comparison on different parameters in POT and UGRN}
Table~\ref{table:para-pot} reports
how different parameters impact the performance and the complexity of our model. The results show that, enlarging the embedding dimension from 48 to 96 can boost the performance, but using dimensions larger than 96 cannot bring further benefits. In addition, we observe the best performance when using 12 and 3 transformer layers in POT encoder and UGRN, respectively, and no more gains can be obtained by stacking more layers. Therefore, we set the basic setting to $L_1 =12$, $L_2 =3$, and $C=96$.

\subsubsection{Comparison on model complexity}
In Table~\ref{tab:model_complexity}, We compare both the accuracy and the model complexity with other benchmarks on the Human3.6M dataset. We provide two configurations of our method, in which the embedding dimension of Our-S is 48 while that of Our-L is set to 96.
Results show that our method can achieve better results with even much fewer parameters.

\subsubsection{Understanding the performance improvement}
In Figure~\ref{fig:Hard joints}, we present the average estimation errors of different body joints according to its group index.  It can be seen that, both our group position embedding and UGRN bring more performance improvement for group 4 and 5, in which joints are far from the root joint. The results confirm that our benefit mainly comes form the difficult joints.

\subsubsection{Qualitative results}
Figure~\ref{fig:vis} demonstrates some qualitative results on the Human3.6M dataset compared with Graformer~\cite{zhao2022graformer}. It can be seen that our method can make accurate pose prediction, especially for the difficult joints that are far from the root.

\section{Conclusion} 
In this paper, we proposed a two-stage transformer-based framework for 3D HPE. First, we introduce targeted improvements for the basic components of transformers and fabricate Pose-Oriented Transformer (POT). Specifically, we design a novel self-attention mechanism in which the topological connections of body joints can be well considered. We also split body joints into several groups according to their distance toward the root joint and provide additional learnable distance-related position embedding for each group. Then, the second stage Uncertainty-Guided Refinement Network (UGRN) is introduced to further refine pose predictions, by considering the estimated uncertainty of each joint with uncertainty-guided sampling strategy and self-attention mechanism. Extensive results on Human3.6M and MPI-INF-3DHP reveal the benefits of our design.
\section{Acknowledgments}
This work was supported in part by the National Natural Science Foundation of China under Grant 61932022, Grant 61931023, Grant 61971285, Grant 61831018, Grant 61871267, Grant 61720106001, Grant 62120106007, Grant 61972256, Grant T2122024, Grant 62125109, and in part by the Program of Shanghai Science and Technology Innovation Project under Grant 20511100100.    

\bibliography{aaai23}

\begin{thebibliography}{39}
\providecommand{\natexlab}[1]{#1}

\bibitem[{Cai et~al.(2019)Cai, Ge, Liu, Cai, Cham, Yuan, and
  Thalmann}]{2019Exploiting}
Cai, Y.; Ge, L.; Liu, J.; Cai, J.; Cham, T.-J.; Yuan, J.; and Thalmann, N.~M.
  2019.
\newblock Exploiting Spatial-Temporal Relationships for {3D} Pose Estimation
  via Graph Convolutional Networks.
\newblock In \emph{2019 IEEE/CVF International Conference on Computer Vision
  (ICCV)}, 2272--2281.

\bibitem[{Chen et~al.(2018)Chen, Wang, Peng, Zhang, Yu, and
  Sun}]{chen2018cascaded}
Chen, Y.; Wang, Z.; Peng, Y.; Zhang, Z.; Yu, G.; and Sun, J. 2018.
\newblock Cascaded pyramid network for multi-person pose estimation.
\newblock In \emph{2018 IEEE/CVF Conference on Computer Vision and Pattern
  Recognition}, 7103--7112.

\bibitem[{Ci et~al.(2019)Ci, Wang, Ma, and Wang}]{ci2019optimizing}
Ci, H.; Wang, C.; Ma, X.; and Wang, Y. 2019.
\newblock Optimizing network structure for {3D} human pose estimation.
\newblock In \emph{2019 IEEE/CVF International Conference on Computer Vision
  (ICCV)}, 2262--2271.

\bibitem[{Dosovitskiy et~al.(2020)Dosovitskiy, Beyer, Kolesnikov, Weissenborn,
  Zhai, Unterthiner, Dehghani, Minderer, Heigold, Gelly
  et~al.}]{dosovitskiy2020image}
Dosovitskiy, A.; Beyer, L.; Kolesnikov, A.; Weissenborn, D.; Zhai, X.;
  Unterthiner, T.; Dehghani, M.; Minderer, M.; Heigold, G.; Gelly, S.; et~al.
  2020.
\newblock An image is worth 16x16 words: Transformers for image recognition at
  scale.
\newblock \emph{arXiv preprint arXiv:2010.11929}.

\bibitem[{Gal and Ghahramani(2016)}]{gal2016dropout}
Gal, Y.; and Ghahramani, Z. 2016.
\newblock Dropout as a bayesian approximation: Representing model uncertainty
  in deep learning.
\newblock In \emph{international conference on machine learning}, 1050--1059.
  PMLR.

\bibitem[{Glorot and Bengio(2010)}]{glorot2010understanding}
Glorot, X.; and Bengio, Y. 2010.
\newblock Understanding the difficulty of training deep feedforward neural
  networks.
\newblock In \emph{Proceedings of the 13th International Conference on
  Artificial Intelligence and Statistics (AISTATS) 2010}, 249--256.

\bibitem[{Ionescu et~al.(2013)Ionescu, Papava, Olaru, and
  Sminchisescu}]{ionescu2013human3}
Ionescu, C.; Papava, D.; Olaru, V.; and Sminchisescu, C. 2013.
\newblock {Human3.6M}: Large scale datasets and predictive methods for {3D}
  human sensing in natural environments.
\newblock \emph{IEEE Transactions on Pattern Analysis and Machine
  Intelligence}, 36(7): 1325--1339.

\bibitem[{Kendall and Gal(2017)}]{kendall2017uncertainties}
Kendall, A.; and Gal, Y. 2017.
\newblock What uncertainties do we need in bayesian deep learning for computer
  vision?
\newblock \emph{Advances in neural information processing systems}, 30.

\bibitem[{Kingma and Ba(2014)}]{kingma2014adam}
Kingma, D.~P.; and Ba, J. 2014.
\newblock Adam: A method for stochastic optimization.
\newblock In \emph{2nd International Conference on Learning Representations}.

\bibitem[{Li et~al.(2021)Li, Shi, Dai, Chen, Wang, Sun, Guo, Li, Zou, and
  Xiong}]{li2021hierarchical}
Li, H.; Shi, B.; Dai, W.; Chen, Y.; Wang, B.; Sun, Y.; Guo, M.; Li, C.; Zou,
  J.; and Xiong, H. 2021.
\newblock Hierarchical Graph Networks for 3D Human Pose Estimation.
\newblock \emph{The British Machine Vision Conference}.

\bibitem[{Lin and Lee(2020)}]{lin2020hdnet}
Lin, J.; and Lee, G.~H. 2020.
\newblock {HDNet}: Human Depth Estimation for Multi-Person Camera-Space
  Localization.
\newblock In \emph{Proceedings of the European Conference on Computer Vision},
  633--648.

\bibitem[{Lin, Wang, and Liu(2021)}]{lin2021mesh}
Lin, K.; Wang, L.; and Liu, Z. 2021.
\newblock Mesh graphormer.
\newblock In \emph{Proceedings of the IEEE/CVF International Conference on
  Computer Vision}, 12939--12948.

\bibitem[{Liu et~al.(2020)Liu, Ding, Zou, Wang, and
  Tang}]{liu2020comprehensive}
Liu, K.; Ding, R.; Zou, Z.; Wang, L.; and Tang, W. 2020.
\newblock A comprehensive study of weight sharing in graph networks for {3D}
  human pose estimation.
\newblock In \emph{Proceedings of the European Conference on Computer Vision},
  318--334.

\bibitem[{Luo, Chu, and Yuille(2020)}]{luo2018orinet}
Luo, C.; Chu, X.; and Yuille, A. 2020.
\newblock Orinet: A fully convolutional network for 3d human pose estimation.
\newblock In \emph{British Machine Vision Conference}.

\bibitem[{Martinez et~al.(2017)Martinez, Hossain, Romero, and
  Little}]{2017simple}
Martinez, J.; Hossain, R.; Romero, J.; and Little, J.~J. 2017.
\newblock A simple yet effective baseline for {3D} human pose estimation.
\newblock In \emph{2017 IEEE International Conference on Computer Vision
  (ICCV)}, 2640--2649.

\bibitem[{Mehta et~al.(2017)Mehta, Rhodin, Casas, Fua, Sotnychenko, Xu, and
  Theobalt}]{mehta2017monocular}
Mehta, D.; Rhodin, H.; Casas, D.; Fua, P.; Sotnychenko, O.; Xu, W.; and
  Theobalt, C. 2017.
\newblock Monocular 3d human pose estimation in the wild using improved cnn
  supervision.
\newblock In \emph{international conference on 3D vision}, 506--516. IEEE.

\bibitem[{Moon, Chang, and Lee(2019)}]{moon2019camera}
Moon, G.; Chang, J.~Y.; and Lee, K.~M. 2019.
\newblock Camera distance-aware top-down approach for {3D} multi-person pose
  estimation from a single {RGB} image.
\newblock In \emph{2019 IEEE/CVF International Conference on Computer Vision},
  10133--10142.

\bibitem[{Newell, Yang, and Deng(2016)}]{newell2016stacked}
Newell, A.; Yang, K.; and Deng, J. 2016.
\newblock Stacked hourglass networks for human pose estimation.
\newblock In \emph{Proceedings of the European Conference on Computer Vision},
  483--499.

\bibitem[{Pavlakos, Zhou, and Daniilidis(2018)}]{2018Ordinal}
Pavlakos, G.; Zhou, X.; and Daniilidis, K. 2018.
\newblock Ordinal depth supervision for {3D} human pose estimation.
\newblock In \emph{2018 IEEE/CVF Conference on Computer Vision and Pattern
  Recognition}, 7307--7316.

\bibitem[{Pavllo et~al.(2019)Pavllo, Feichtenhofer, Grangier, and
  Auli}]{pavllo20193d}
Pavllo, D.; Feichtenhofer, C.; Grangier, D.; and Auli, M. 2019.
\newblock {3D} human pose estimation in video with temporal convolutions and
  semi-supervised training.
\newblock In \emph{2019 IEEE/CVF Conference on Computer Vision and Pattern
  Recognition (CVPR)}, 7753--7762.

\bibitem[{Shi et~al.(2020)Shi, Xu, Dai, Wang, Zhang, Li, Zou, and
  Xiong}]{shi2020tiny}
Shi, B.; Xu, Y.; Dai, W.; Wang, B.; Zhang, S.; Li, C.; Zou, J.; and Xiong, H.
  2020.
\newblock {Tiny-Hourglassnet}: An Efficient Design For {3D} Human Pose
  Estimation.
\newblock In \emph{2020 IEEE International Conference on Image Processing
  (ICIP)}, 1491--1495.

\bibitem[{Song et~al.(2021)Song, Chen, Zheng, and Ji}]{song2021uncertain}
Song, T.; Chen, L.; Zheng, W.; and Ji, Q. 2021.
\newblock Uncertain graph neural networks for facial action unit detection.
\newblock In \emph{Proceedings of the AAAI Conference on Artificial
  Intelligence}, volume~35, 5993--6001.

\bibitem[{Sun et~al.(2017)Sun, Shang, Liang, and Wei}]{sun2017compositional}
Sun, X.; Shang, J.; Liang, S.; and Wei, Y. 2017.
\newblock Compositional human pose regression.
\newblock In \emph{2017 IEEE International Conference on Computer Vision
  (ICCV)}, 2602--2611.

\bibitem[{Vaswani et~al.(2017)Vaswani, Shazeer, Parmar, Uszkoreit, Jones,
  Gomez, Kaiser, and Polosukhin}]{vaswani2017attention}
Vaswani, A.; Shazeer, N.; Parmar, N.; Uszkoreit, J.; Jones, L.; Gomez, A.~N.;
  Kaiser, {\L}.; and Polosukhin, I. 2017.
\newblock Attention is all you need.
\newblock \emph{Advances in neural information processing systems}, 30.

\bibitem[{Wang et~al.(2021)Wang, Li, Guo, Fang, and Wang}]{wang2021data}
Wang, Z.; Li, Y.; Guo, Y.; Fang, L.; and Wang, S. 2021.
\newblock Data-uncertainty guided multi-phase learning for semi-supervised
  object detection.
\newblock In \emph{Proceedings of the IEEE/CVF Conference on Computer Vision
  and Pattern Recognition}, 4568--4577.

\bibitem[{Wehrbein et~al.(2021)Wehrbein, Rudolph, Rosenhahn, and
  Wandt}]{wehrbein2021probabilistic}
Wehrbein, T.; Rudolph, M.; Rosenhahn, B.; and Wandt, B. 2021.
\newblock Probabilistic monocular 3d human pose estimation with normalizing
  flows.
\newblock In \emph{Proceedings of the IEEE/CVF international conference on
  computer vision}, 11199--11208.

\bibitem[{Xu and Takano(2021)}]{2021Graph}
Xu, T.; and Takano, W. 2021.
\newblock Graph Stacked Hourglass Networks for {3D} Human Pose Estimation.
\newblock In \emph{2021 IEEE/CVF Conference on Computer Vision and Pattern
  Recognition (CVPR)}, 16105--16114.

\bibitem[{Yang et~al.(2021)Yang, Zhai, Li, Huang, Luo, Cheng, and
  Fan}]{yang2021uncertainty}
Yang, F.; Zhai, Q.; Li, X.; Huang, R.; Luo, A.; Cheng, H.; and Fan, D.-P. 2021.
\newblock Uncertainty-guided transformer reasoning for camouflaged object
  detection.
\newblock In \emph{Proceedings of the IEEE/CVF International Conference on
  Computer Vision}, 4146--4155.

\bibitem[{Yang et~al.(2018)Yang, Ouyang, Wang, Ren, Li, and Wang}]{yang20183d}
Yang, W.; Ouyang, W.; Wang, X.; Ren, J.; Li, H.; and Wang, X. 2018.
\newblock 3d human pose estimation in the wild by adversarial learning.
\newblock In \emph{2018 IEEE Conference on Computer Vision and Pattern
  Recognition (CVPR)}, 5255--5264.

\bibitem[{Ying et~al.(2021)Ying, Cai, Luo, Zheng, Ke, He, Shen, and
  Liu}]{ying2021transformers}
Ying, C.; Cai, T.; Luo, S.; Zheng, S.; Ke, G.; He, D.; Shen, Y.; and Liu, T.-Y.
  2021.
\newblock Do transformers really perform badly for graph representation?
\newblock \emph{Advances in Neural Information Processing Systems}, 34:
  28877--28888.

\bibitem[{Zhang et~al.(2022)Zhang, Tu, Yang, Chen, and Yuan}]{zhang2022mixste}
Zhang, J.; Tu, Z.; Yang, J.; Chen, Y.; and Yuan, J. 2022.
\newblock MixSTE: Seq2seq Mixed Spatio-Temporal Encoder for 3D Human Pose
  Estimation in Video.
\newblock In \emph{Proceedings of the IEEE/CVF Conference on Computer Vision
  and Pattern Recognition}, 13232--13242.

\bibitem[{Zhao et~al.(2019)Zhao, Peng, Tian, Kapadia, and
  Metaxas}]{zhao2019semantic}
Zhao, L.; Peng, X.; Tian, Y.; Kapadia, M.; and Metaxas, D.~N. 2019.
\newblock Semantic graph convolutional networks for {3D} human pose regression.
\newblock In \emph{2019 IEEE/CVF Conference on Computer Vision and Pattern
  Recognition (CVPR)}, 3425--3435.

\bibitem[{Zhao, Wang, and Tian(2022)}]{zhao2022graformer}
Zhao, W.; Wang, W.; and Tian, Y. 2022.
\newblock GraFormer: Graph-Oriented Transformer for 3D Pose Estimation.
\newblock In \emph{Proceedings of the IEEE/CVF Conference on Computer Vision
  and Pattern Recognition}, 20438--20447.

\bibitem[{Zheng et~al.(2021)Zheng, Zhu, Mendieta, Yang, Chen, and
  Ding}]{zheng20213d}
Zheng, C.; Zhu, S.; Mendieta, M.; Yang, T.; Chen, C.; and Ding, Z. 2021.
\newblock 3d human pose estimation with spatial and temporal transformers.
\newblock In \emph{Proceedings of the IEEE/CVF International Conference on
  Computer Vision}, 11656--11665.

\bibitem[{Zhou et~al.(2019)Zhou, Han, Jiang, Jia, and Lu}]{zhou2019hemlets}
Zhou, K.; Han, X.; Jiang, N.; Jia, K.; and Lu, J. 2019.
\newblock Hemlets pose: Learning part-centric heatmap triplets for accurate 3d
  human pose estimation.
\newblock In \emph{2017 IEEE International Conference on Computer Vision
  (ICCV)}, 2344--2353.

\bibitem[{Zhou et~al.(2017)Zhou, Huang, Sun, Xue, and Wei}]{zhou2017towards}
Zhou, X.; Huang, Q.; Sun, X.; Xue, X.; and Wei, Y. 2017.
\newblock Towards {3D} human pose estimation in the wild: A weakly-supervised
  approach.
\newblock In \emph{2017 IEEE International Conference on Computer Vision
  (ICCV)}, 398--407.

\bibitem[{Zhu et~al.(2021)Zhu, Xu, Shen, Ji, Gao, and Shen}]{zhu2021posegtac}
Zhu, Y.; Xu, X.; Shen, F.; Ji, Y.; Gao, L.; and Shen, H.~T. 2021.
\newblock PoseGTAC: Graph Transformer Encoder-Decoder with Atrous Convolution
  for 3D Human Pose Estimation.
\newblock In \emph{IJCAI}, 1359--1365.

\bibitem[{Zou et~al.(2020)Zou, Liu, Wang, and Tang}]{zou2020high}
Zou, Z.; Liu, K.; Wang, L.; and Tang, W. 2020.
\newblock High-order Graph Convolutional Networks for {3D} Human Pose
  Estimation.
\newblock In \emph{British Machine Vision Conference}.

\bibitem[{Zou and Tang(2021)}]{zou2021modulated}
Zou, Z.; and Tang, W. 2021.
\newblock Modulated graph convolutional network for 3d human pose estimation.
\newblock In \emph{Proceedings of the IEEE/CVF International Conference on
  Computer Vision}, 11477--11487.

\end{thebibliography}

\end{document}